\documentclass[12pt]{article}%
\usepackage{amsmath}
\usepackage{amsfonts}
\usepackage{amssymb}
\usepackage{graphicx}%
\providecommand{\U}[1]{\protect\rule{.1in}{.1in}}

\begin{document}

\title{A weighted random survival forest}
\author{Lev V. Utkin$^{1}$, Andrei V. Konstantinov$^{1}$, Viacheslav S. Chukanov$^{1}$,
\and Mikhail V. Kots$^{1}$, Mikhail A. Ryabinin$^{1}$, Anna A. Meldo$^{1,2}$\\$^{1}${\small Peter the Great St.Petersburg Polytechnic University (SPbPU)}\\$^{2}${\small St.Petersburg Clinical Research Center for Special Types of
Medical Care (Oncology-oriented)}}
\date{}
\maketitle

\begin{abstract}
A weighted random survival forest is presented in the paper. It can be
regarded as a modification of the random forest improving its performance. The
main idea underlying the proposed model is to replace the standard procedure
of averaging used for estimation of the random survival forest hazard function
by weighted avaraging where the weights are assigned to every tree and can be
veiwed as training paremeters which are computed in an optimal way by solving
a standard quadratic optimization problem maximizing Harrell's C-index.
Numerical examples with real data illustrate the outperformance of the proposed
model in comparison with the original random survival forest.


\end{abstract}

\section{Introduction}

A lot of computer aided diagnosis (CAD) systems have been developed in order
to provide successful detection of a disease and to facilitate making decision
to start treatment process at early stage. Most CAD systems aim to detect a
disease or its features. However, there are a few systems which take into
account survival aspects of a patient especially of a cancer patient. Two
reasons of such the situation can be pointed out. First, CAD systems taking
into account survival aspects require the corresponding datasets which are
mainly lack or of a small size nowadays. Second, CAD systems have to handle
data with censored observations. This peculiarity may significantly complicate
training process, and it requires special methods for dealing with censored
data. A large amount of structured data, which has been recorded about
patients, their peculiarities, do not take into account survival aspects.
Therefore, it is topical to develop models which could efficiently process the
available survival datasets in order to be an element of CAD systems.

A basis for such the models may be survival analysis or time-to-event analysis
which can be regarded as a fundamental tool which is used in many applied
areas. One of the most important areas is the medical research where survival
models are widely used to evaluate the significance of prognostic variables in
outcomes such as death or cancer recurrence and subsequently inform patients
of their treatment options \cite{Katzman-etal-2018}. The datasets used in the
survival analysis or just the survival data differ from many datasets by the
fact that time to event of interest for a part of observations or instances is
unknown because the event might not have happened during the period of study
\cite{Nezhad-etal-2018}. If the observed survival time is less than or equal
to the true survival time, then we have a special case of censoring data
called right-censoring data. Other special cases are left-censoring and
interval censoring observations \cite{Wang-Li-Reddy-2017}. However,
right-censoring is the most common case in many applications
\cite{Hosmer-Lemeshow-May-2008}. Without loss of generality, we describe the
survival models mainly in the medical application terms below, i.e., instances
will be called patients.

The survival models can be divided into three parts: parametric, nonparametric
and semiparametric. It is assumed in parametric models that the type of the
probability distribution of survival times is known, for example, the
exponential, Weibull, normal, gamma distributions. As pointed out by Lee and
Wang \cite{Lee-Wang-2003}, nonparametric or distribution-free models are less
efficient than parametric methods when survival times follow a theoretical
distribution and more efficient when no suitable theoretical distributions are
known. They can be used to analyze survival data before attempting to fit a
theoretical distribution. One of the simplest survivor models is the
Kaplan-Meier estimator which is a non-parametric model used to compute the
survival function of a homogeneous data set. In other words, the model does
not take into account the fact that the instances may differ by their
features. A few critical features of the Kaplan-Meier model are considered in
\cite[Chapter 4, Page 76]{Lee-Wang-2003}. Nevertheless, the Kaplan-Meier model
provides a simple way to compute the survival function of patients.

A popular regression model for the analysis of survival data is the well-known
Cox proportional hazards model, which is a semi-parametric model that
calculates the effects of observed covariates on the risk of an event
occurring, for example, the death or failure \cite{Cox-1972}. The Cox model is
the most commonly used regression analysis approach for survival data among
semi-parametric survival models. The model does not require knowledge of the
underlying distribution. It differs significantly from other methods since it
is built on the proportional hazards assumption and employs partial likelihood
for parameter estimation. The proportional hazards assumption in the Cox model
means that different patients have hazard functions that are proportional,
i.e., the ratio of the hazard functions for two patients with different
prognostic factors or covariates is a constant and does not vary with time. In
other words, the ratio of the risk of dying of two patients is the same no
matter how long they survive \cite[Chapter 12]{Lee-Wang-2003}. The model
assumes that a patient's log-risk of failure is a linear combination of the
patient's covariates. This assumption is referred to as the linear
proportional hazards condition. It is interesting to note that the Cox model
is semi-parametric in the sense that it can be factored into a parametric part
consisting of a regression parameter vector associated with the covariates and
a non-parametric part that can be left completely unspecified
\cite{Devarajn-Ebrahimi-2011}. Another interpretation of the semi-parametric
property of the Cox model is that we do not require to know the underlying
distribution of time to event of interest, but the attributes are assumed to
have an exponential influence on the outcome \cite{Wang-Li-Reddy-2017}.

The Cox model is a very powerful method for dealing with survival data. As a
result, a lot of approaches dealing with the Cox model and its modifications
have been proposed last decades. A clear taxonomy of survival analysis methods
and their comprehensive review are presented by Wang et al.
\cite{Wang-Li-Reddy-2017}.

It should be noted that the Cox model may provide unsatisfactory results under
conditions of a high dimensionality of survivor data and a small number of
observations. These conditions take place in many application problems, for
example, when we deal with gene expression data. However, due to the high
dimensionality of gene expression data when the number of genes expressed
exceeds the number of patients, it is not possible to take an estimation
approach based on the standard Cox model. To overcome this problem, Tibshirani
\cite{Tibshirani-1997} proposed one of the interesting modifications of the
Cox model based on the Lasso method. Kim et al. \cite{Kim-etal-2012}
considered the Cox regression with the group Lasso penalty which improves the
combination of different covariates, for example, clinical and genomic
covariates. The adaptive Lasso for the Cox model is proposed by
\cite{Zhang-Lu-2007}. Some modifications of the Cox model with using Lasso can
also be found in
\cite{Fan-Feng-Wu-2010,Kaneko-etal-2015,Krasotkina-Mottl-2015,Ternes-etal-2016,Witten-Tibshirani-2010}%
.

One of the main problems of the Cox model is linear relationship assumption
between covariates and the time of event occurrence. Various modifications
have been proposed to generalize the Cox model taking into account the
corresponding non-linear relationship between covariates and the time of
event. The first class of models uses a neural network for modelling the
non-linear function. Faraggi and Simon in their pioneering work
\cite{Faraggi-Simon-1995} presented an approach to modelling survival data
using the input-output relationship associated with a simple feed-forward
neural network as the basis for a non-linear proportional hazards model. The
proposed model was a basis for developing more complex generalization using
the deep neural networks
\cite{Katzman-etal-2018,Luck-etal-2017,Nezhad-etal-2018,Ranganath-etal-2016,Yao-Zhu-Zhu-Huang-2017,Zhu-Yao-Huang-2016}%
. The convolutional neural networks (CNN) also have been applied to the
survival analysis. In particular, Haarburger et al.
\cite{Haarburger-etal-2018} used CNN for analysis of lung cancer patients and
illustrated that the CNN improves the predictive accuracy of Cox models that
otherwise only rely on radiomics features. Some aspects of application of the
survival analysis to medical diagnostic problems have been discussed by Afshar
at al. \cite{Afshar-etal-2018}. Several models based on neural networks are
considered in the review by Wang et al. \cite{Wang-Li-Reddy-2017}. A review of
deep learning methods for dealing with survival data is presented by Nezhad et
al. \cite{Nezhad-etal-2018}. The proposed generalizations have many
advantages, but there is an important disadvantage. The use of neural networks
requires a lot of survival data. This condition is violated in many
applications. Therefore, Van Belle et al.
\cite{Van_Belle-etal-2007,Van_Belle-etal-2008} proposed to use SVM in order to
enhance the model by the small amount of training data. The SVM approach to
survival analysis has been studied by several authors
\cite{Khan-Zubek-2008,Polsterl-etal-2016,Shivaswamy-etal-2007,Van_Belle-etal-2011,Widodo-Yang-2011}%
.

Another approach for dealing with the limited survival data is to use survival
trees and the random survival forests (RSFs). As pointed out by Wang et al.
\cite{Wang-Li-Reddy-2017}, the splitting criteria as one of the main concepts
of decision trees differ for survival trees. The splitting criteria can be
divided into two classes: minimizing within-node homogeneity and maximizing
between-node heterogeneity. The first class of approaches minimizes the loss
function using the within-node homogeneity criterion. Criteria from the first
class measure the within-node homogeneity with a statistic that measures how
similar the subjects in each node are and choose splits that minimize the
within-node error. In particular, Gordon and Olsen \cite{Gordon-Olshen-1985}
proposed an extension of CART to survival data by applying a distance measure,
for example, the Wasserstein metric, between Kaplan-Meier curves and certain
point masses. Davis and Anderson \cite{Davis-Anderson-1989} proposed another
splitting criterion based on the likelihood method under assumption that the
survivor function in a node is exponential with a constant hazard. An example
of a splitting criterion from the second class is a criterion using the
log-rank test statistics presented by Ciampi \cite{Ciampi-1991}. Due to many
advantages of decision trees as a tool for classification and regression,
several tree-based modifications solving the survival analysis problem have
been proposed last decades
\cite{Huang-Chen-Soong-1998,Ibrahim-etal-2008,LeBlanc-Crowley-1992,Linden-Yarnold-2017,Segal-1988,Su-Fan-2004,Yoon-etal-2018,Zhang-1995}%
. Survival random forests have been applied to many real application problems,
for example, \cite{Akai-etal-2018,Gilhodes-etal-2017,Miao-etal-2015}. A
detailed review of survival trees as well as RSFs is represented by Bou-Hamad
et al. \cite{Bou-Hamad-etal-2011}. A new algorithm for rule induction from
survival data was proposed by Wrobel et al. \cite{Wrobel-etal-2017}. It works
according to the separate-and-conquer heuristics with a use of log-rank test
for establishing rule body.

Random forests were introduced by Breiman \cite{Breiman-2001} in order to
overcome some shortcomings of the decision trees, in particular, their
instability to small perturbations in a learning sample. The random forest
uses a large number of randomly built patient decision trees in order to
combine their predictions. It also reduces the possible correlation between
decision trees by selecting different subsamples of the feature space.

It turns out that the random forests became a very powerful, efficient and
popular tool for the survival analysis. The random forest can be regarded as a
nonparametric machine learning strategy. The popularity of RSFs stems from
many useful factors. First of all, Ishwaran and Kogalur
\cite{Ishwaran-Kogalur-2007} pointed out that the random forests require only
three tuning parameters to be set (the number of randomly selected predictors,
the number of trees grown in the forest, and the splitting rule). Moreover,
the random forest is highly data adaptive and virtually model assumption free.
Wang and Zhou \cite{Wang-Zhou-2017} mention also that random forests have
proved to be successful in various scenarios including classification,
regression and survival analysis \cite{Biau-Scornet-2016}. They can deal with
both low and high dimensional data while other popular ensembles often fail
when confronted with high dimensional datasets. As a result, a lot of models
based on random forest have been developed for dealing with survival data
\cite{Bou-Hamad-etal-2011a,Hu-Steingrimsson-2018,Ishwaran-etal-2004,Khalilia-etal-2011,Mogensen-etal-2012,Nasejje-etal-2017,Omurlu-etal-2009,Schmid-etal-2016,Taylor-2011,Wright-etal-2016,Wright-etal-2017}%
. Most models are very similar and differ in splitting criteria and the
ensemble estimation. Splitting criteria totally defines the survival trees in
the random forest and has been briefly considered above. Most survival random
forests use averaging of the tree cumulative hazard estimates and its modifications.

It should be noted that other ensemble models and algorithms for dealing with
survival data have been developed, for example, Hothorn et al.
\cite{Hothorn-etal-2006} proposed a unified and flexible framework for
ensemble survival learning and introduced the corresponding random forest and
generic gradient boosting algorithms.

Since the RSF is one of the most efficient models in survival analysis, then
we pay attention to this model and propose an approach for its improving. The
first idea underlying the improvement is to modify the procedure of averaging
used for estimation of the forest survival function on the basis of survival
functions of trees. We propose to replace the standard averaging with the
weighted sum of the tree survival functions. The corresponding RSF with
weights will be called weighted RSF (WRSF). By assigning the weights to every
tree survival function, we, in fact, assign these weights to every tree in the
random forest because the weights do not depend on the training examples. The
second idea is that weights in the sum are regarded as training parameters
which can be computed in an optimal way by solving an optimization problem.
The third idea is to apply the concordance error rate called C-index
\cite{Harrell-etal-1982} for constructing the optimization problem. The
C-index estimates how good the model is at ranking survival times. It is one
of the popular measures for comparison survival models. It turns out that
maximization of the C-index may be a basis for training the tree weights. It
should be noted that the use of the C-index in its original form makes the
optimization problem computationally hard to be solved. Therefore, the fourth
idea is to replace the C-index with its approximate representation which is
based on applying the well-known hinge loss function. As a result, we get the
standard quadratic optimization problem for computing optimal weights, which
can be solved by many available methods.

The weighting scheme in random forests is not new. Some random forest
algorithms assign weights to classes \cite{Daho-2014}. There are algorithms
with weights of decision trees
\cite{Kim-Kim-Moon-Ahn-2011,Li-Wang-Ding-Dong-2010,Ronao-Cho-2015}. However,
to the best of our knowledge, the weighting schemes have not been used in
RSFs. Moreover, in contrast to the available weighting algorithms in original
random forests, the proposed approach considers weights in the RSF as training parameters.

\section{Some elements of survival analysis and a formal problem statement}

In survival analysis, a patient $i$ is represented by a triplet $(\mathbf{x}%
_{i},\delta_{i},T_{i})$, where $\mathbf{x}_{i}=(x_{i1},...,x_{im})$ is the
vector of the patient parameters (characteristics) or the vector of features;
$T_{i}$ indicates time to event of the patient, it is assumed to be
non-negative and continuous. If the event of interest is observed, $T_{i}$
corresponds to the time between baseline time and the time of event happening,
in this case $\delta_{i}=1$, and we have an uncensored observation. If the
instance event is not observed and its time to event is greater than the
observation time, $T_{i}$ corresponds to the time between baseline time and
end of the observation, and the event indicator is $\delta_{i}=0$, and we have
a censored observation. Suppose a training set $D$ consists of $n$ triplets
$(\mathbf{x}_{i},\delta_{i},T_{i})$, $i=1,...,n$. The goal of survival
analysis is to estimate the time to the event of interest $T$ for a new
patient with feature vector denoted by $\mathbf{x}$ by using the training set
$D$.

The survival and hazard functions are key concepts in survival analysis for
describing the distribution of event times. The survival function denoted by
$S(t)$ as a function of time $t$ is the probability of surviving up to that
time, i.e., $S(t)=\Pr\{T>t\}$. The hazard function $h(t)$ is the rate of event
at time $t$ given that no event occurred before time $t$, i.e.,
\[
h(t)=\lim_{\Delta t\rightarrow0}\frac{\Pr\{t\leq T\leq t+\Delta t|T\geq
t\}}{\Delta t}=\frac{f(t)}{S(t)},
\]
where $f(t)$ is the density function of the event of interest.

By using the fact that the density function can be expressed through the
survival function as
\[
f(t)=-\frac{dS(t)}{dt},
\]
we can write the following expression for the hazard rate:
\[
h(t)=-\frac{d}{dt}\ln S(t).
\]

The survival function is determined through the hazard function as
\[
S(t)=\exp\left(  -\int_{0}^{t}h(z)\mathrm{d}z\right)  =\exp\left(
-H(t)\right)  ,
\]
where $H(t)$ is the cumulative hazard function.

We did not write the dependence of the above functions on a feature vector
$\mathbf{x}$ for short.

\subsection{The Cox model}

According to the Cox proportional hazards model,
\cite{Hosmer-Lemeshow-May-2008}, the hazard function at time $t$ given
predictor values $\mathbf{x}$ is defined as
\[
h(t|\mathbf{x})=h_{0}(t)\Psi(\mathbf{x},\mathbf{b})=h_{0}(t)\exp\left(
\psi(\mathbf{x},\mathbf{b})\right)  .
\]

Here $h_{0}(t)$ is an arbitrary baseline hazard function; $\Psi(\mathbf{x})$
is the covariate effect or the risk function; $\mathbf{b}=(b_{1},...,b_{m})$
is an unknown vector of regression coefficients or parameters. It can be seen
from the above expression for the hazard function that the reparametrization
$\Psi(\mathbf{x},\mathbf{b})=\exp\left(  \psi(\mathbf{x},\mathbf{b})\right)  $
is used in the Cox model. The function $\psi(\mathbf{x},\mathbf{b})$ in the
model is linear, i.e.,
\[
\psi(\mathbf{x},\mathbf{b})=\mathbf{xb}^{\mathrm{T}}=\sum\nolimits_{k=1}%
^{m}b_{k}x_{k}.
\]

In the framework of the Cox model, the survival function $S(t)$ is computed
as
\[
S(t)=\exp(-H_{0}(t)\exp\left(  \psi(\mathbf{x},\mathbf{b})\right)  )=\left(
S_{0}(t)\right)  ^{\exp\left(  \psi(\mathbf{x},\mathbf{b})\right)  }.
\]

Here $H_{0}(t)$ is the cumulative baseline hazard function; $S_{0}(t)$ is the
baseline survival function.

The partial likelihood in this case is defined as follows:
\[
L(\mathbf{b})=\prod_{j=1}^{n}\left[  \frac{\exp(\psi(\mathbf{x}_{j}%
,\mathbf{b}))}{\sum_{i\in R_{j}}\exp(\psi(\mathbf{x}_{i},\mathbf{b}))}\right]
^{\delta_{j}}.
\]

Here $R_{j}$ is the set of patients who are at risk at time $t_{j}$. The term
\textquotedblleft at risk at time $t$\textquotedblright\ means patients who
die at time $t$ or later.

It should be note that the idea underlying the use of neural networks in
survival analysis is to replace the linear function $\psi(\mathbf{x})$ with a
non-linear function which is realized by means of a neural network
\cite{Faraggi-Simon-1995}.

In order to provide personalized treatment recommendations in accordance with
the recommender function, we compute the functions $\psi_{i}(\mathbf{x})$ and
$\psi_{j}(\mathbf{x})$ corresponding to different treatment groups. If the
obtained function $rec_{ij}(\mathbf{x})$ is positive, then treatment $j$ is
preferable in comparison with treatment $i$. In the case of a negative
recommender function, treatment $i$ is more effective and leads to a lower
risk than treatment $j$ (see, for example, \cite{Katzman-etal-2018}).

To compare the survival models, the C-index proposed by Harrell et al.
\cite{Harrell-etal-1982} is used. The C-index estimates how good the model is
at ranking survival times. It estimates the probability that, in a randomly
selected pair of patients, the patient that fails first had a worst predicted
outcome. In fact, this is the probability that the event times of a pair of
patients are correctly ranking. C-index does not depend on choosing a fixed
time for evaluation of the model and takes into account censoring of patients
\cite{May-etal-2004}.

Let us consider the training set $D$ consisting of $n$ triplets $(\mathbf{x}%
_{i},\delta_{i},T_{i})$. We consider possible or admissible pairs
$\{(\mathbf{x}_{i},\delta_{i},T_{i}),(\mathbf{x}_{j},\delta_{j},T_{j})\}$ for
$i\leq j$. Then the C-index is calculated as the ration of the number of pairs
correctly ordered by the model to the total number of admissible pairs. A pair
is not admissible if the events are both right-censored or if the earliest
time in the pair is censored. If the C-index is equal to 1, then the
corresponding survival model is supposed to be perfect. If the C-index is 0.5,
then the model is no better than random guessing.

Let $t_{1}^{\ast},...,t_{q}^{\ast}$ denote predefined time points, for
example, $t_{1},...,t_{N}$, where $N$ is distinct event times. If the output
of a survival algorithm is the predicted survival function $S(t)$, then the
C-index is formally calculated as \cite{Wang-Li-Reddy-2017}:%
\begin{equation}
C=\frac{1}{M}\sum_{i:\delta_{i}=1}\sum_{j:t_{i}<t_{j}}\mathbf{1}\left[
S(t_{i}^{\ast}|\mathbf{x}_{i})>S(t_{j}^{\ast}|\mathbf{x}_{j})\right]  .
\label{Survival_DF_24}%
\end{equation}

Here $M$ is the number of all comparable or admissible pairs; $\mathbf{1}[a]$
is the indicator function taking the value $1$ if $a$ is true, and $0$
otherwise; $S$ is the estimated survival function.

It should be noted that there are different definitions of the C-index, which
depend on the output of a survival algorithm. However, we will use the
definition (\ref{Survival_DF_24}) which plays an important role in the
proposed improvement of the RSF.

\subsection{Random survival forests}

It has been mentioned that the RSF is one of the best models for survival
analysis due to its properties. This is the main reason for its modifying
below in order to improve the survival analysis results and to increase the
prediction accuracy.

A general algorithm of constructing RSFs can be represented as follows
\cite{Ishwaran-etal-2008}:

1. Draw $Q$ bootstrap samples from the original data. Note that each bootstrap
sample excludes on average 37\% of the data, called out-of-bag data (OOB data).

2. Grow a survival tree for each bootstrap sample. At each node of the tree,
randomly select $\sqrt{m}$ candidate variables. The node is split using the
candidate variable that maximizes survival difference between daughter nodes.

3. Grow the tree to full size under the constraint that a terminal node should
have no less than $d>0$ unique deaths.

4. Calculate a cumulative hazard function for each tree or a survival
function. Average to obtain the ensemble cumulative hazard function or the
ensemble survival function.

5. Using out-of-bag data, calculate prediction error for the ensemble
cumulative hazard function or the ensemble survival function.

The parameters of the algorithm proposed by Ishwaran et al.
\cite{Ishwaran-etal-2008} and some its steps may vary, but, generally, it can
be viewed as a basis for solving the survival analysis problem by means of
many its implementations and modifications.

The most important question of the RSFs, which defines their different
implementations is the splitting rule. As shown by Ishwaran et al.
\cite{Ishwaran-etal-2008}, a good split maximizes survival difference across
the two sets of data. We shortly review the main splitting rules used in RSF
\cite{Ishwaran-etal-2008,Wang-Li-Reddy-2017}.

Let $t_{1}<t_{2}<...<t_{N}$ be the distinct times to event of interest, for
example, times to deaths, in the parent node $g$, and let $Z_{ij}$ and
$Y_{ij}$ equal the number of deaths and patients at risk at time $t_{i}$ in
the daughter nodes $j=1,2$, i.e.,
\[
Y_{i1}=\#\{T_{l}\geq t_{i},\ x_{l}\leq c\},\ Y_{i2}=\#\{T_{l}\geq
t_{i},\ x_{l}>c\}.
\]

Here $x_{l}$ is the value of a feature $x$ for the $l$-th patient,
$l=1,...,n$. Let $Y_{i}=Y_{i1}+Y_{i2}$ and $Z_{i}=Z_{i1}+Z_{i2}$. Let $n_{1}$
and $n_{2}$ be total numbers of observations in daughter nodes such that
$n=n_{1}+n_{2}$, i.e.,%
\[
n_{1}=\#\{l:\ x_{l}\leq c\},\ n_{2}=\#\{l:\ x_{l}>c\}.
\]

The \textit{log-rank test} for a split at the value $c$ for predictor $x$ is
defined as%
\[
L(x,c)=\frac{\sum_{i=1}^{N}\left(  Z_{i1}-Y_{i1}Z_{i}/Y_{i}\right)  }%
{\sqrt{\sum_{i=1}^{N}\frac{Y_{i1}}{Y_{i}}\left(  1-\frac{Y_{i1}}{Y_{i}%
}\right)  \left(  \frac{Y_{i}-Z_{i}}{Y_{i}-1}\right)  Z_{i}}}.
\]

The value $\left\vert L(x,c)\right\vert $ is the measure of node separation,
which should be minimized for better splitting.

An idea underlying another splitting rule called as \textit{conservation of
events splitting} is to suppose that the sum of estimated cumulative hazard
functions over the observed time points must equal the total number of deaths.
By using the notations introduced for the log-rank test, the measure of
conservation of events for the split on $x$ at the value $c$ can be defined
as
\[
Cons(x,c)=\frac{1}{Y_{11}+Y_{12}}\sum_{j=1}^{2}Y_{1j}\sum_{k=1}^{N-1}\left\{
N_{kj}Y_{k+1,j}\sum_{l=1}^{k}\frac{Z_{lj}}{Y_{lj}}\right\}  .
\]

It should be noted that the splitting rule should maximize survival
differences due to the split. Therefore, the transformed value
$1/(1+Cons(x,c))$ as a measure of node separation is used.

We also consider the \textit{approximate log-rank splitting}. Let
$Z=\sum_{i=1}^{N}Z_{i}$ and $Z_{1}=\sum_{i=1}^{N}Z_{i1}$. The log-rank test
$L(x,c)$ is%
\[
L(x,c)=\frac{Z^{1/2}\left(  Z_{1}-\sum_{l=1}^{n}\mathbf{1}\left\{  x_{l}\leq
c\right\}  H(T_{l})\right)  }{\sqrt{\left(  \sum_{l=1}^{n}\mathbf{1}\left\{
x_{l}\leq c\right\}  H(T_{l})\right)  \left(  Z-\sum_{l=1}^{n}\mathbf{1}%
\left\{  x_{l}\leq c\right\}  H(T_{l})\right)  }}.
\]

The next important question is how to compute the ensemble hazard function or
the ensemble survival function. First, we consider how to compute the
cumulative hazard estimate for the $k$-th terminal node of a tree. Let
$\{t_{j,k}\}$ be the $N(k)$ distinct death times in terminal node $k$ of the
$q$-th tree such that $t_{1,k}<t_{2,k}<...<t_{N(k),k}$ and $Z_{j,k}$ and
$Y_{j,k}$ equal the number of deaths and patients at risk at time $t_{j,k}$.
The cumulative hazard estimate for node $k$ is defined as (the Nelson--Aalen
estimator):
\[
H_{k}(t)=\sum_{t_{j,k}\leq t}Z_{j,k}/Y_{j,k}.
\]

If the $i$-th patient with features $\mathbf{x}_{i}$ falls into node $k$, then
we can say that $H(t|\mathbf{x}_{i})=H_{k}(t)$. The ensemble cumulative hazard
estimate for the $i$-th patient is obtained by averaging cumulative hazard
estimates of all $Q$ trees, i.e.,
\begin{equation}
H_{f}(t|\mathbf{x}_{i})=\frac{1}{Q}\sum_{q=1}^{Q}H_{q}(t|\mathbf{x}_{i}).
\label{Survival_DF_36}%
\end{equation}

The survival function can be obtained from $H_{q}(t|\mathbf{x}_{i})$ as
follows:
\[
S_{q}(t|\mathbf{x}_{i})=\exp\left(  -H_{q}(t|\mathbf{x}_{i})\right)  .
\]

Another ensemble estimate is considered by Ishwaran et al.
\cite{Ishwaran-etal-2008}, where OOB data are used. Let $O_{q}$ be a set of
OBB example indexes for the tree $q$. The OOB prediction for each training
example $\mathbf{x}_{i}$ uses only the trees that did not have $\mathbf{x}%
_{i}$ in their bootstrap sample. If we define the indicator function as
$\mathbf{1}(i\in O_{q})$, then the OOB ensemble cumulative hazard estimator
for the $i$-th training example is defined as
\begin{equation}
H_{f}(t|\mathbf{x}_{i})=\frac{\sum_{q=1}^{Q}\mathbf{1}(i\in O_{q})\cdot
H_{q}(t|\mathbf{x}_{i})}{\sum_{q=1}^{Q}\mathbf{1}(i\in O_{q})}.
\label{Survival_DF_37}%
\end{equation}

\section{Weights of survival decision trees}

One can see from (\ref{Survival_DF_36}) that the ensemble cumulative hazard
estimate $H_{f}(t|\mathbf{x}_{i})$ is obtained under condition that all trees
have the same weights $1/Q$. A straightforward way to improve the random
forest is to assign weights $\mathbf{w}=(w_{1},...,w_{Q})$ to decision trees.
At that, it is assumed that the sum of weight is $1$, i.e., every vector
$\mathbf{w}$ belongs to the unit simplex of the dimension $Q$. As a result, we
replace the averaging of the cumulative hazard estimates (\ref{Survival_DF_36}%
) by weighted averaging for computing the the cumulative hazard function as
follows:
\begin{equation}
H_{f}(t,\mathbf{w}|\mathbf{x}_{i})=\sum_{q=1}^{Q}w_{q}H_{q}(t|\mathbf{x}_{i}).
\label{Survival_DF_38}%
\end{equation}
One of the ways for assigning the weights is to suppose that they are training
parameters which can be optimized in accordance with a goal. Therefore, we
have to define the goal or an objective function for getting optimal weights.

One of the most important measure for comparison different models is the
C-index defined in (\ref{Survival_DF_24}). If we assume that the predicted
survival function of the random forest depends on the weights, we can maximize
the C-index with respect to the weights. Let us write the C-index as a
function of the weights%
\begin{equation}
C(\mathbf{w})=\frac{1}{M}\sum_{i:\delta_{i}=1}\sum_{j:t_{i}<t_{j}}%
\mathbf{1}\left[  S_{f}(t_{i}^{\ast},\mathbf{w}|\mathbf{x}_{i})-S_{f}%
(t_{j}^{\ast},\mathbf{w}|\mathbf{x}_{j})>0\right]  \mathbf{.}
\label{Survival_DF_39}%
\end{equation}

Here $S_{f}(t_{i}^{\ast},\mathbf{w}|\mathbf{x}_{i})$ is the ensemble predicted
survival function depending on weights $\mathbf{w}$ of trees. By maximizing
the $C(\mathbf{w})$ over the non-negative weights $w_{q}$, $q=1,...,Q$, under
constraint $\sum_{q=1}^{Q}w_{q}=1$, we can get optimal weights.

It is difficult to solve the optimization problem with the indicator function
in the objective function (\ref{Survival_DF_39}) because we have a hard
combinatorial problem. Moreover, the dependence of the ensemble survivor
function on the weights is non-linear because
\[
S_{f}(t,\mathbf{w}|\mathbf{x}_{i})=\exp\left(  -H_{f}(t,\mathbf{w}%
|\mathbf{x}_{i})\right)  =\exp\left(  -\sum_{q=1}^{Q}w_{q}H_{q}(t|\mathbf{x}%
_{i})\right)  .
\]

Fortunately, we can overcome this difficulty as follows. Note that there holds%
\begin{align*}
&  \mathbf{1}\left[  S_{f}(t_{i}^{\ast},\mathbf{w}|\mathbf{x}_{i})>S_{f}%
(t_{j}^{\ast},\mathbf{w}|\mathbf{x}_{j})\right] \\
&  =\mathbf{1}\left[  \ln S_{f}(t_{i}^{\ast},\mathbf{w}|\mathbf{x}_{i})>\ln
S_{f}(t_{j}^{\ast},\mathbf{w}|\mathbf{x}_{j})\right] \\
&  =\mathbf{1}\left[  H_{f}(t_{j}^{\ast},\mathbf{w}|\mathbf{x}_{i}%
)>H_{f}(t_{i}^{\ast},\mathbf{w}|\mathbf{x}_{j})\right]  .
\end{align*}
Hence, we get
\[
C(\mathbf{w})=\frac{1}{M}\sum_{i:\delta_{i}=1}\sum_{j:t_{i}<t_{j}}%
\mathbf{1}\left[  H_{f}(t_{j}^{\ast},\mathbf{w}|\mathbf{x}_{i})-H_{f}%
(t_{i}^{\ast},\mathbf{w}|\mathbf{x}_{j})>0\right]  \mathbf{.}%
\]

Let us denote the set of all possible pairs $(i,j)$ in (\ref{Survival_DF_39}),
satisfying the condition $\delta_{i}=1$ for $i$ and the condition $t_{i}%
<t_{j}$ for $j$, as $J$. Taking into account (\ref{Survival_DF_38}), we get
the following optimization problem:
\begin{equation}
C(\mathbf{w})=\max_{\mathbf{w}}\frac{1}{M}\sum_{(i,j)\in J}\mathbf{1}\left[
\sum_{q=1}^{Q}w_{q}\left(  H_{q}(t_{j}^{\ast}|\mathbf{x}_{j})-H_{q}%
(t_{i}^{\ast}|\mathbf{x}_{i})\right)  >0\right]  , \label{Survival_DF_40}%
\end{equation}
subject to
\begin{equation}
\sum_{q=1}^{Q}w_{q}=1,\ w_{q}\geq0,\ q=1,...,Q. \label{Survival_DF_44}%
\end{equation}

The constraints for weights produce the unit simplex denoted as $\Delta_{Q}$
whose dimensionality is $Q$. By maximizing $C(\mathbf{w})$ over $\mathbf{w}%
\in\Delta_{Q}$, we can get optimal weights.

One of the obvious ways for simplifying the optimization problem is to replace
the indicator function with the sigmoid $\sigma$, i.e., the optimization
problem becomes to be%
\begin{equation}
C(\mathbf{w})=\max_{\mathbf{w}\in\Delta_{Q}}\frac{1}{M}\sum_{(i,j)\in J}%
\sigma\left[  \sum_{q=1}^{Q}w_{q}\left(  H_{q}(t_{j}^{\ast}|\mathbf{x}%
_{j})-H_{q}(t_{i}^{\ast}|\mathbf{x}_{i})\right)  \right]  .
\label{Survival_DF_42}%
\end{equation}

It can be seen from the objective function that the problem can be solved by
applying the gradient descent method. However, the main difficulty here is to
take into account the linear constraints for weights (\ref{Survival_DF_44})
which can be represented as the unit simplex of weights denoted as $\Delta
_{Q}$ whose dimensionality is $Q$.

Another way for simplifying the optimization problem is to replace the
indicator function with the hinge loss function similarly to the replacement
proposed by Van Belle et al. \cite{Van_Belle-etal-2007}. The hinge loss
function is of the form:
\[
l(x)=\max\left(  0,x\right)  .
\]

By adding the regularization term, we can write the optimization problem as%
\begin{equation}
\min_{\mathbf{w}\in\Delta_{Q}}\left\{  \sum_{(i,j)\in J}\max\left(
0,\sum_{q=1}^{Q}w_{q}\left(  H_{q}(t_{i}^{\ast}|\mathbf{x}_{i})-H_{q}%
(t_{j}^{\ast}|\mathbf{x}_{j})\right)  \right)  +\lambda R(\mathbf{w})\right\}
. \label{Survival_DF_46}%
\end{equation}

Here $R(\mathbf{w})$ is a regularization term, $\lambda$ is a hyper-parameter
which controls the strength of the regularization. We define the
regularization term as
\[
R(\mathbf{w})=\left\Vert \mathbf{w}\right\Vert ^{2}.
\]

Let us introduce the variables%
\begin{equation}
\xi_{ij}=\max\left(  0,\sum_{q=1}^{Q}w_{q}\left(  H_{q}(t_{i}^{\ast
}|\mathbf{x}_{i})-H_{q}(t_{j}^{\ast}|\mathbf{x}_{j})\right)  \right)  .
\label{Survival_DF_48}%
\end{equation}
Then the optimization problem is of the form:
\begin{equation}
\min_{\mathbf{w}}\left\{  \sum_{(i,j)\in J}\xi_{ij}+\lambda\left\Vert
\mathbf{w}\right\Vert ^{2}\right\}  , \label{Survival_DF_50}%
\end{equation}
subject to $\mathbf{w}\in\Delta_{Q}$ and
\begin{equation}
\xi_{ij}\geq\sum_{q=1}^{Q}w_{q}\left(  H_{q}(t_{i}^{\ast}|\mathbf{x}%
_{i})-H_{q}(t_{j}^{\ast}|\mathbf{x}_{j})\right)  ,\ \ \xi_{ij}\geq
0,\ \ \{i,j\}\in J. \label{Survival_DF_51}%
\end{equation}

We get a standard quadratic optimization problem with linear constraints and
with the vector $\mathbf{w}$ of $Q$ variables. It can be solved by many known methods.

It is interesting to note that the above optimization problem is very similar
to the primal form of the well-known SVM \cite{Scholkopf-Smola02}.

A general algorithm for training the WRSF taking into account weighted
ensemble estimation can be regarded as an extension of the algorithm given in
previous section for the original RSF. Given the training set $D=\{(\mathbf{x}%
_{i},\delta_{i},T_{i}),\ i=1,...,n\}$, $\mathbf{x}_{i}\in\mathbb{R}^{m}$,
$\delta_{i}\in\{0,1\}$, $T_{i}\in\mathbb{R}$, we use the cumulative hazard
functions $H_{q}(t_{i}^{\ast}|\mathbf{x}_{i})$ of all trees ($q=1,...,Q$)
corresponding to the $i$-th example, $i=1,...,n$, and solve the optimization
problem (\ref{Survival_DF_50})-(\ref{Survival_DF_51}). Taking the optimal
weights $\mathbf{w}$ as the solution of (\ref{Survival_DF_50}%
)-(\ref{Survival_DF_51}), we use (\ref{Survival_DF_38}) in order to get the
ensemble survival function.

It is interesting to note that the above optimization problem is very similar
to the primal form of the SVM modification for survival analysis \cite[Problem
(11)]{Van_Belle-etal-2007}. Indeed, the objective functions are identical.
Constraints in the survival SVM are of the form: $\xi_{ij}\geq1+\sum_{k=1}%
^{m}w_{q}\left(  x_{ik}-x_{jk}\right)  $. One can see that the idea of the SVM
modification for survival analysis is to find a line which separates ranking
points $\mathbf{x}_{i}-\mathbf{x}_{j}$. By using the problem
(\ref{Survival_DF_50})-(\ref{Survival_DF_51}), we try to find a line which
separates the ranking points $\mathbf{H}(t_{i}^{\ast}|\mathbf{x}%
_{i})-\mathbf{H}(t_{j}^{\ast}|\mathbf{x}_{j})$, where $\mathbf{H}%
(t|\mathbf{x})$ is the vector of the cumulative hazard function estimates for
all trees at time $t$ by testing $\mathbf{x}$. If the SVM modification for
survival analysis deals with pairs of feature vectors, then the proposed WRSF
analyses pairs of the decision tree outputs. From this point of view, the
proposed procedure for training the weights of trees can be regarded as a
second-order SVM or meta-learner for the RSF.

It should be noted that the number of weights is equal to the number of trees
in the forest. On the one hand, we would like to improve the classification
algorithm by introducing the weights. On the other hand, we get a lot of
training parameters which may lead to overfitting by a small amount of
training data. In order to overcome this difficulty, we propose to reduce the
number of weights by grouping trees into identical subsets and by assigned
weights to the subsets. Suppose that we divide all trees into $G$ subsets such
that every subset consists of $g$ trees, $g\cdot G=Q$. Then we have $G$
weights and the optimization problem (\ref{Survival_DF_50}%
)-(\ref{Survival_DF_51}) can be rewritten as
\begin{equation}
\min_{\mathbf{w}}\left\{  \sum_{(i,j)\in J}\xi_{ij}+\lambda\left\Vert
\mathbf{w}\right\Vert ^{2}\right\}  ,
\end{equation}
subject to $\mathbf{w}\in\Delta_{G}$ and
\begin{equation}
\xi_{ij}\geq\sum_{k=1}^{G}w_{k}\left(  \widetilde{H}_{k}(t_{i}^{\ast
}|\mathbf{x}_{i})-\widetilde{H}_{k}(t_{j}^{\ast}|\mathbf{x}_{j})\right)
,\ \ \xi_{ij}\geq0,\ \ \{i,j\}\in J.
\end{equation}

Here $\widetilde{H}_{k}(t^{\ast}|\mathbf{x})$ is the mean cumulative hazard
function of the $k$-th subset of trees. The parameters $G$ and $g$ can be
regarded as tuning parameters in place of the parameter $Q$.

Another difficulty of solving the optimization problem (\ref{Survival_DF_50}%
)-(\ref{Survival_DF_51}) is a large number of constraints for $\xi_{ij}$
because all admissible pairs of training data with indices from the set $J$
produce them. It is interesting to point out that the same difficulty has been
considered in the SVM modification for survival analysis
\cite{Van_Belle-etal-2008} where a scalable nearest neighbor algorithm was
proposed to reduce computational load without considerable loss of
performance. According to this algorithm, the number of constraints can be
reduced by selecting a set $J_{i}$ of $k$ samples with a survival time nearest
to the survival time of sample $i$. However, we use another approach. In order
to simplify the optimization problem, we propose to reduce the number of
constraints by random selection of $K$ constraints from the whole set of
constraints which is defined by all pairs of indices in the set $J$. Of
course, we may get a non-optimal solution in this case. However, our numerical
experiments have shown that this simplification of the optimization problem
provides better results than the original RSF.

\section{Numerical experiments}

Since the WRSF can be viewed as an improvement of the original RSF, then our
interest in this study is to compare the weighted RSF and the original RSF.

In order to carry out the comparisons, the proposed weighted RSF is tested on
seven real benchmark datasets. A short introduction of the benchmark datasets
are given below.

The \textbf{Primary Biliary Cirrhosis (PBC) Dataset} contains observations of
418 patients with primary biliary cirrhosis of the liver from the Mayo Clinic
trial \cite{Fleming-Harrington-1991}, 257 of whom have censored data. Every
example is characterized by 17 features including age, sex, ascites, hepatom,
spiders, edema, bili and chol, etc. The dataset can be obtained via the
\textquotedblleft randomForestSRC\textquotedblright\ R package.

The \textbf{German Breast Cancer Study Group 2 (GBSG2) Dataset }contains
observations of 686 women \cite{Sauerbrei-Royston-1999}. Every example is
characterized by 10 features, including age of the patients in years,
menopausal status, tumor size, tumor grade, number of positive nodes, hormonal
therapy, progesterone receptor, estrogen receptor, recurrence free survival
time, censoring indicator (0 - censored, 1 - event). The dataset can be
obtained via the \textquotedblleft TH.data\textquotedblright\ R package.

The \textbf{Chronic Myelogenous Leukemia Survival (CML) Dataset} is simulated
according to structure of the data by the German CML Study Group used in
\cite{Hehlmann-etal-1994}. The dataset consists of 507 observations with 7
feature: a factor with 54 levels indicating the study center; a factor with
levels trt1, trt2, trt3 indicating the treatment group; sex (0 = female, 1 =
male); age in years; risk group (0 = low, 1 = medium, 2 = high); censoring
status (FALSE = censored, TRUE = dead); time survival or censoring time in
days. The dataset can be obtained via the \textquotedblleft
multcomp\textquotedblright\ R package (cml).

The \textbf{Bladder Cancer Dataset (BLCD) }\cite{Pagano-Gauvreau-2000}\textbf{
}(Chapter 21) contains data on 86 patients after surgery assigned to placebo
or chemotherapy (thiopeta). Endpoint is time to recurrence in months. Data on
the number of tumors removed at surgery was also collected. The dataset is
available at http://www.stat.rice.edu/\symbol{126}sneeley/STAT553/Datasets/survivaldata.txt.

The \textbf{Lupus Nephritis Dataset (LND)} \cite{Abrahamowicz-etal-1996}
contains data on 87 persons with lupus nephritis. followed for 15+ years after
an initial renal biopsy (the starting point of follow-up). This data set only
contains time to death/censoring, indicator, duration and log(1+duration),
where duration is the duration of untreated disease prior to biopsy. The
dataset is available at http://www.stat.rice.edu/\symbol{126}sneeley/STAT553/Datasets/survivaldata.txt.

The \textbf{Heart Transplant Dataset (HTD) }contains data on 69 patients
receiving heart transplants \cite{Kalbfleisch-Prentice-1980}. This dataset is
available at http://lib.stat.cmu.edu/datasets/stanford.

The \textbf{Veterans' Administration Lung Cancer Study (Veteran) Dataset}
\cite{Kalbfleisch-Prentice-1980} contains data on 137 males with advanced
inoperable lung cancer. The subjects were randomly assigned to either a
standard chemotherapy treatment or a test chemotherapy treatment. Several
additional variables were also measured on the subjects. The dataset can be
obtained via the \textquotedblleft survival\textquotedblright\ R package.

The WRSF uses a software in Python to implement the procedures for computing
optimal weights of trees, the corresponding C-index of the whole random
forest, and other procedures required for training and testing the WRSF. The
software is available at https://github.com/andruekonst/weighted-random-survival-forest.

To evaluate the C-index, we perform a cross-validation with $100$ repetitions,
where in each run, we randomly select 75\% of data for training and 25\% for
testing. Different values for the regularization hyper-parameter $\lambda$
have been tested, choosing those leading to the best results.

Table \ref{t:RSF_WRSF_results} summarizes the numerical results for RSF and
WRSF by different datasets (column 1). At that, Table \ref{t:RSF_WRSF_results}
shows the mean values of the C-index (columns 2 and 3), the standard deviation
(Std) (columns 4 and 5) and the median of the C-index (columns 6 and 7). It
can be seen from Table \ref{t:RSF_WRSF_results} that the WRSF outperforms the
RSF for all datasets. It is also interesting to point out that the standard
deviation is decreased when we use WRSF.%

\begin{table}[tbp] \centering
\caption{Comparison of the RSF and WRSF for different datasets}%
\begin{tabular}
[c]{ccccccc}\hline
& \multicolumn{2}{c}{Mean value} & \multicolumn{2}{c}{Std} &
\multicolumn{2}{c}{Median}\\\hline
Dataset & RSF & WRSF & RSF & WRSF & RSF & WRSF\\\hline
PBC & $0.888$ & $0.910$ & $0.013$ & $0.010$ & $0.889$ & $0.911$\\\hline
GBSG2 & $0.889$ & $0.910$ & $0.013$ & $0.010$ & $0.891$ & $0.911$\\\hline
BLCD & $0.880$ & $0.934$ & $0.060$ & $0.042$ & $0.891$ & $0.942$\\\hline
CML & $0.889$ & $0.910$ & $0.013$ & $0.010$ & $0.889$ & $0.911$\\\hline
LND & $0.882$ & $0.941$ & $0.051$ & $0.043$ & $0.877$ & $0.944$\\\hline
HTD & $0.859$ & $0.931$ & $0.056$ & $0.044$ & $0.873$ & $0.943$\\\hline
Veteran & $0.870$ & $0.929$ & $0.046$ & $0.041$ & $0.882$ & $0.943$\\\hline
\end{tabular}
\label{t:RSF_WRSF_results}%
\end{table}%

\begin{figure}
[ptb]
\begin{center}
\includegraphics[
height=2.5979in,
width=2.8072in
]%
{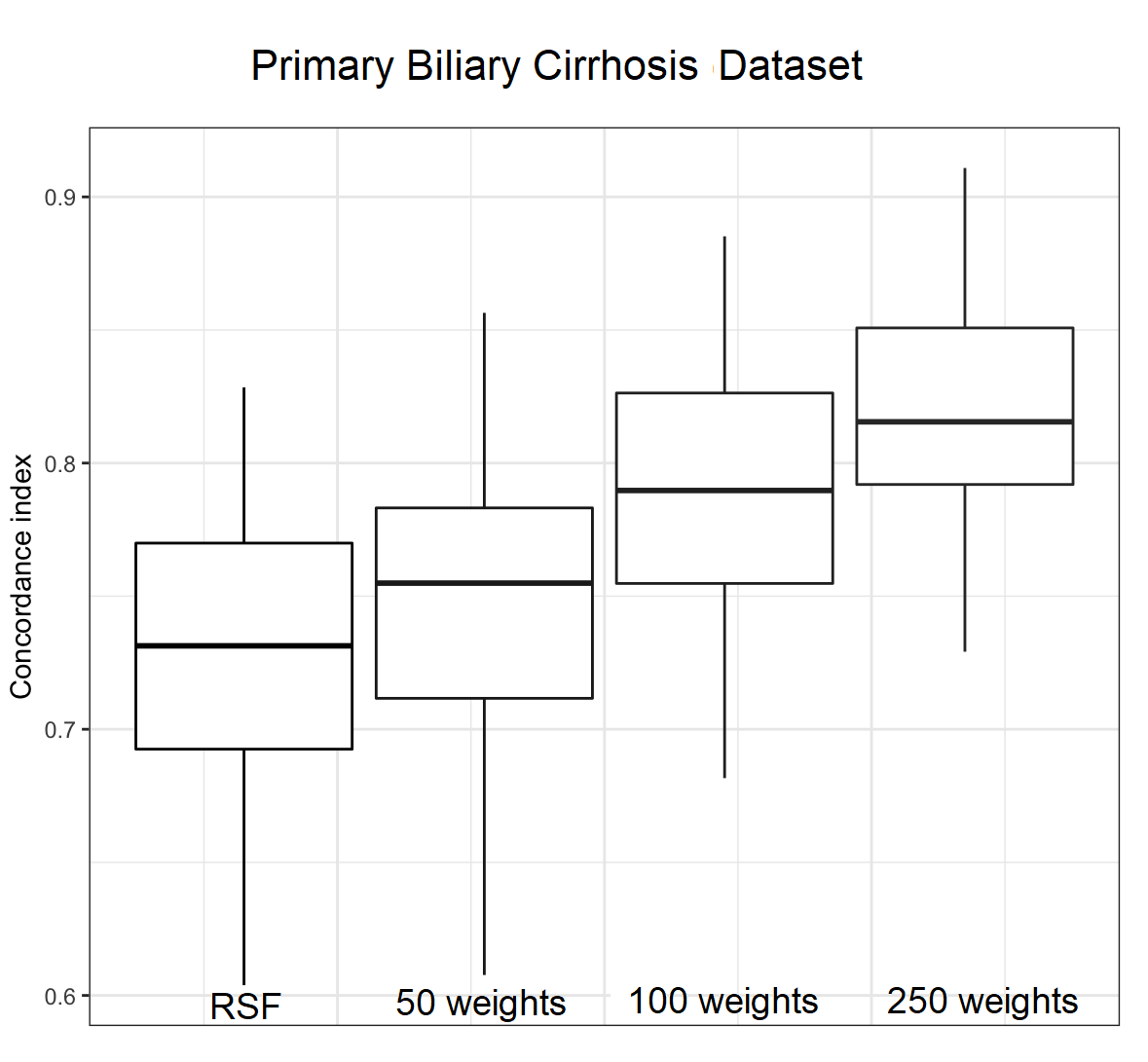}%
\caption{The boxplot for the PBC dataset}%
\label{t:PBC}%
\end{center}
\end{figure}

\begin{figure}
[ptb]
\begin{center}
\includegraphics[
height=2.5348in,
width=2.8383in
]%
{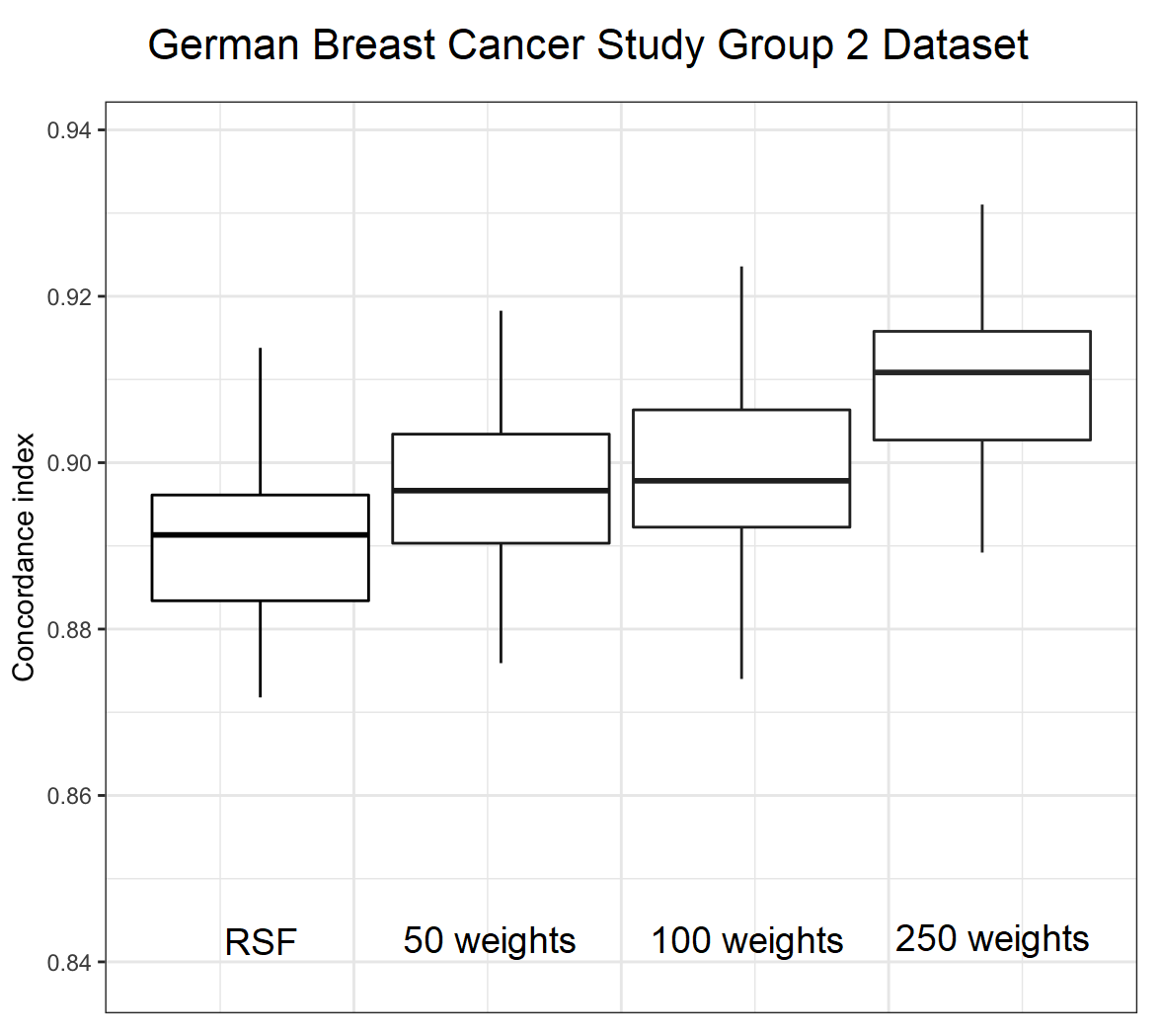}%
\caption{The boxplot for the GCSG2 dataset}%
\label{t:GCSG2}%
\end{center}
\end{figure}

\begin{figure}
[ptb]
\begin{center}
\includegraphics[
height=2.5607in,
width=2.8539in
]%
{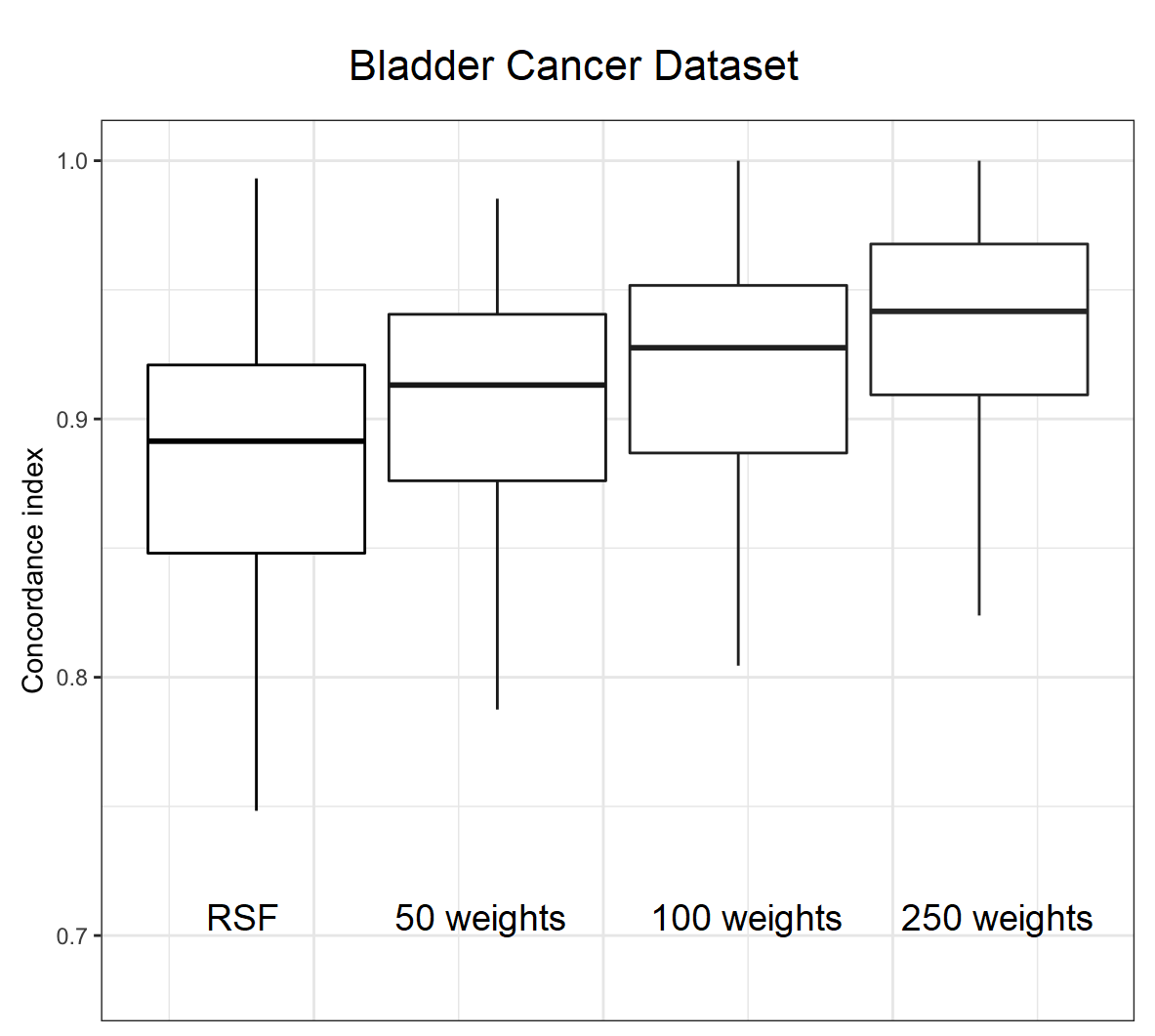}%
\caption{The boxplot for the BLCD dataset}%
\label{t:BLCD}%
\end{center}
\end{figure}

\begin{figure}
[ptb]
\begin{center}
\includegraphics[
height=2.5746in,
width=2.8452in
]%
{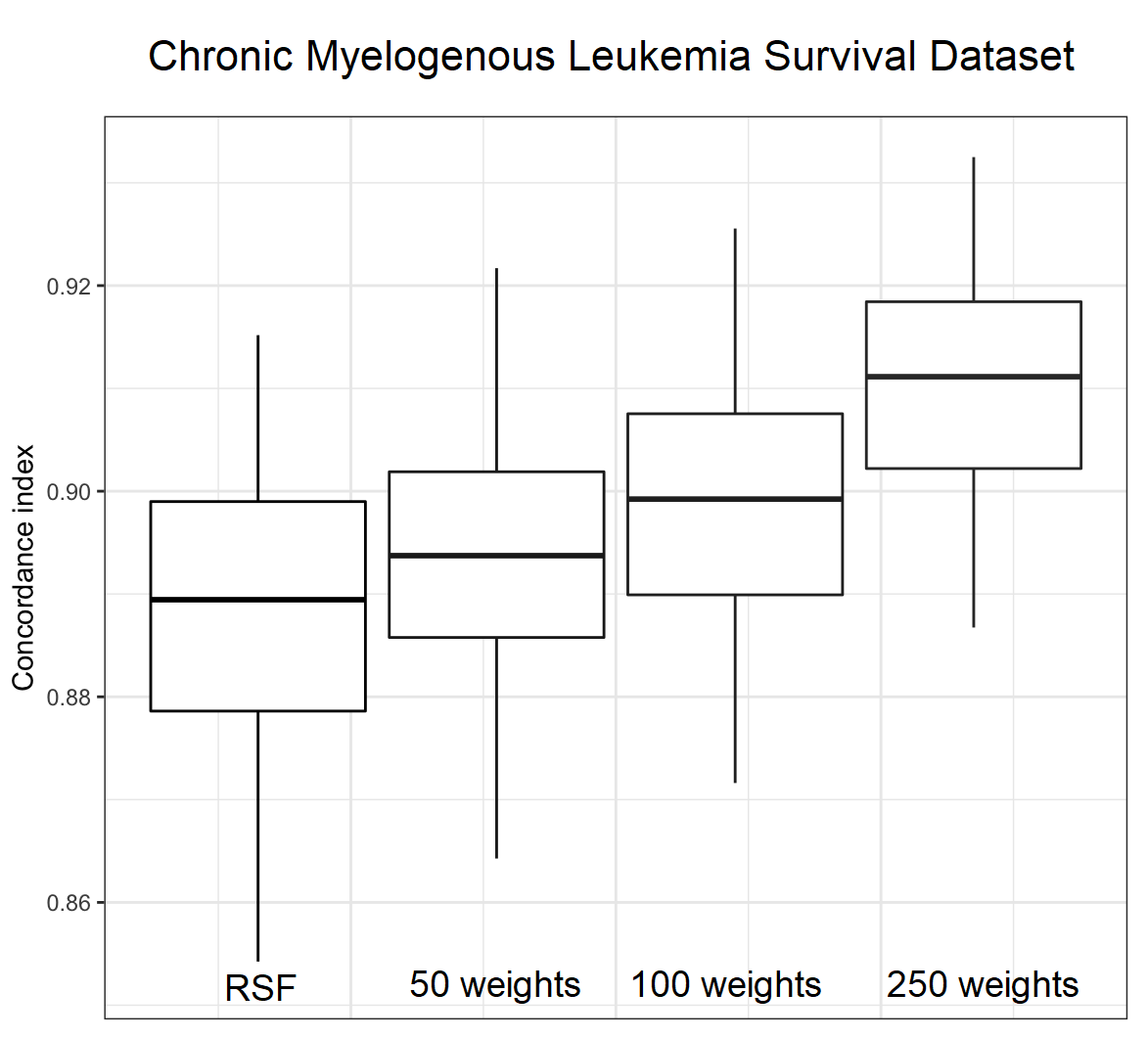}%
\caption{The boxplot for the CML dataset}%
\label{t:CML}%
\end{center}
\end{figure}

\begin{figure}
[ptb]
\begin{center}
\includegraphics[
height=2.5244in,
width=2.8349in
]%
{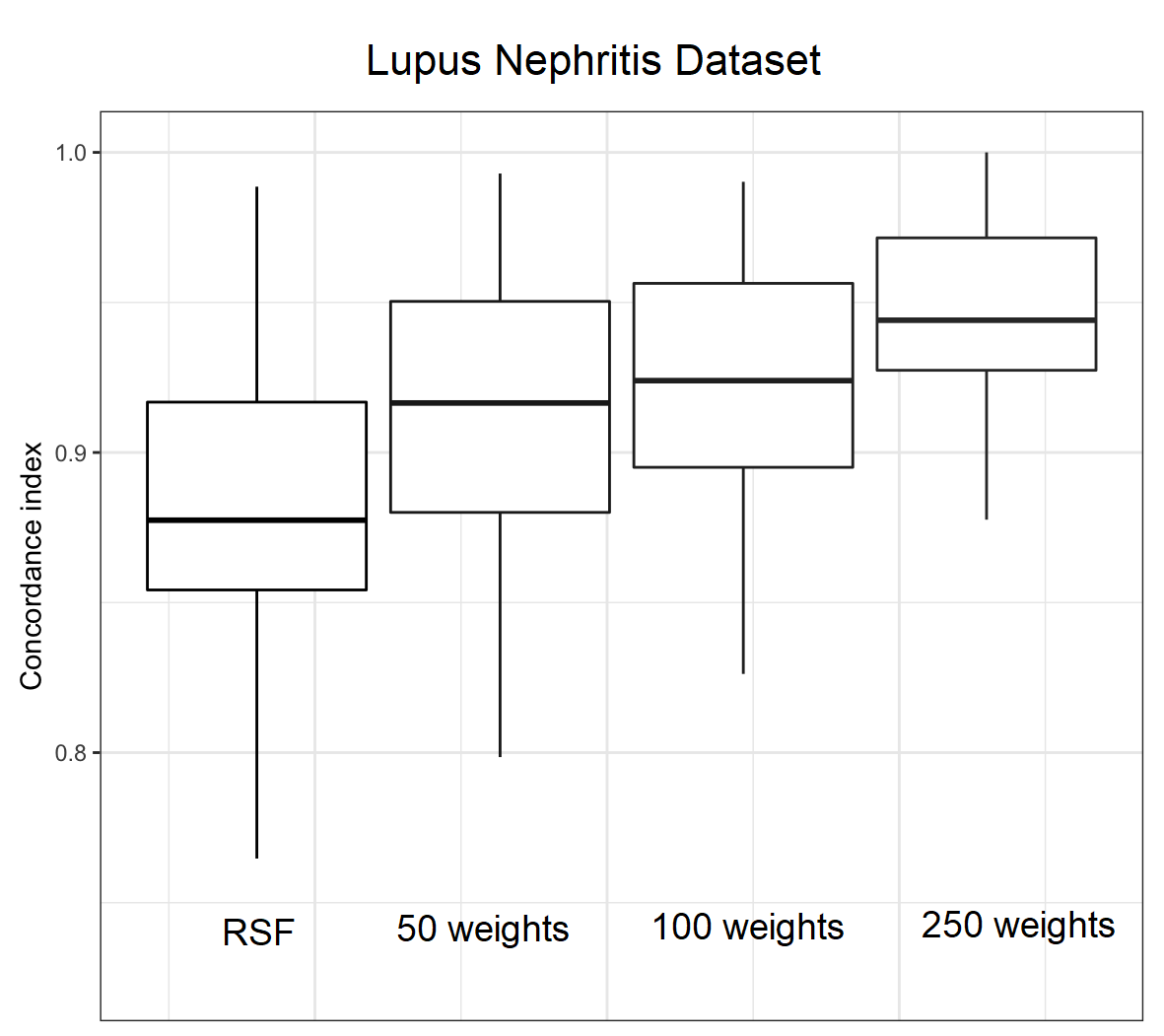}%
\caption{The boxplot for the LND dataset}%
\label{t:LND}%
\end{center}
\end{figure}

\begin{figure}
[ptb]
\begin{center}
\includegraphics[
height=2.5702in,
width=2.8331in
]%
{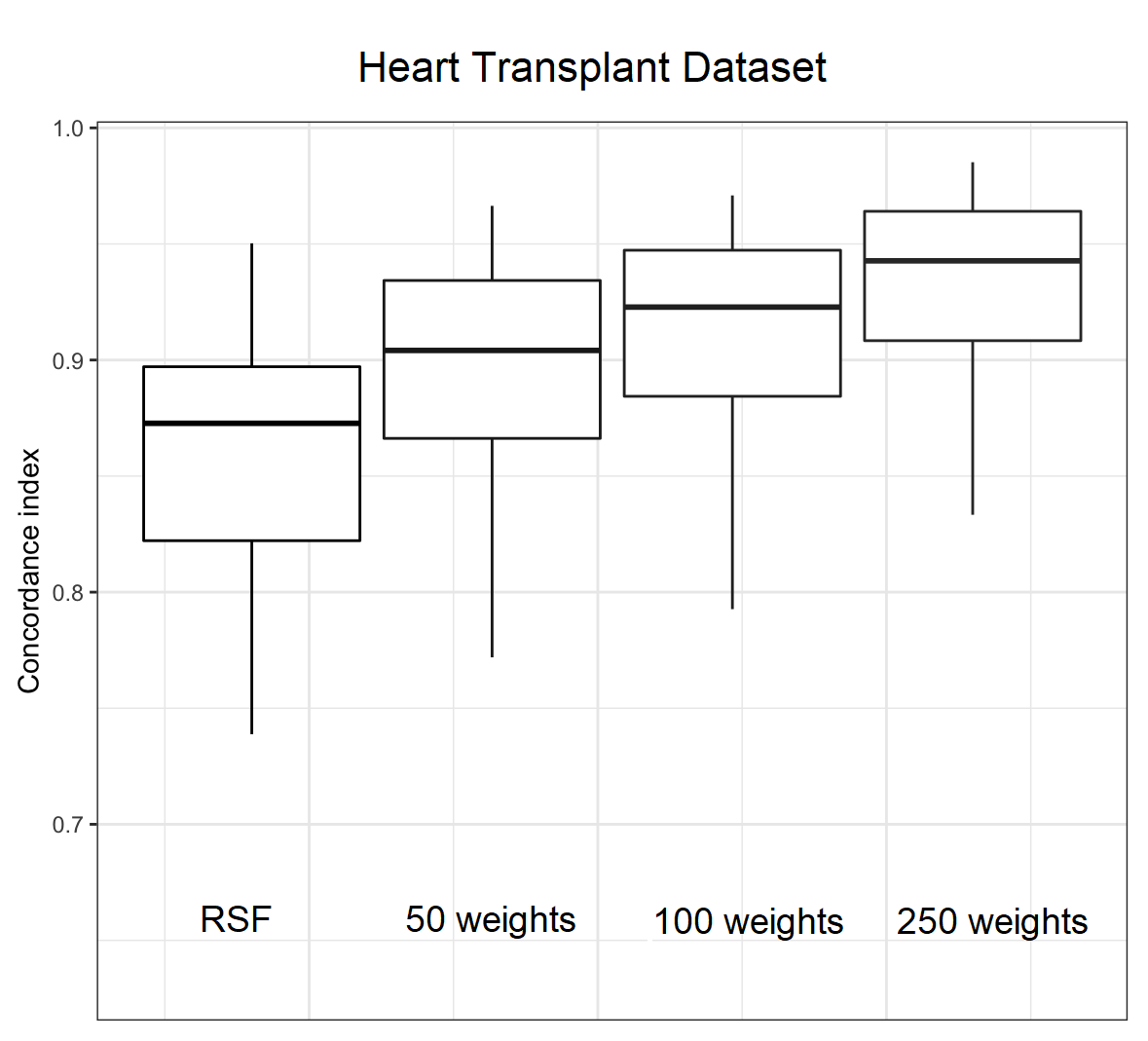}%
\caption{The boxplot for the HTD dataset}%
\label{t:HTD}%
\end{center}
\end{figure}

\begin{figure}
[ptb]
\begin{center}
\includegraphics[
height=2.565in,
width=2.8487in
]%
{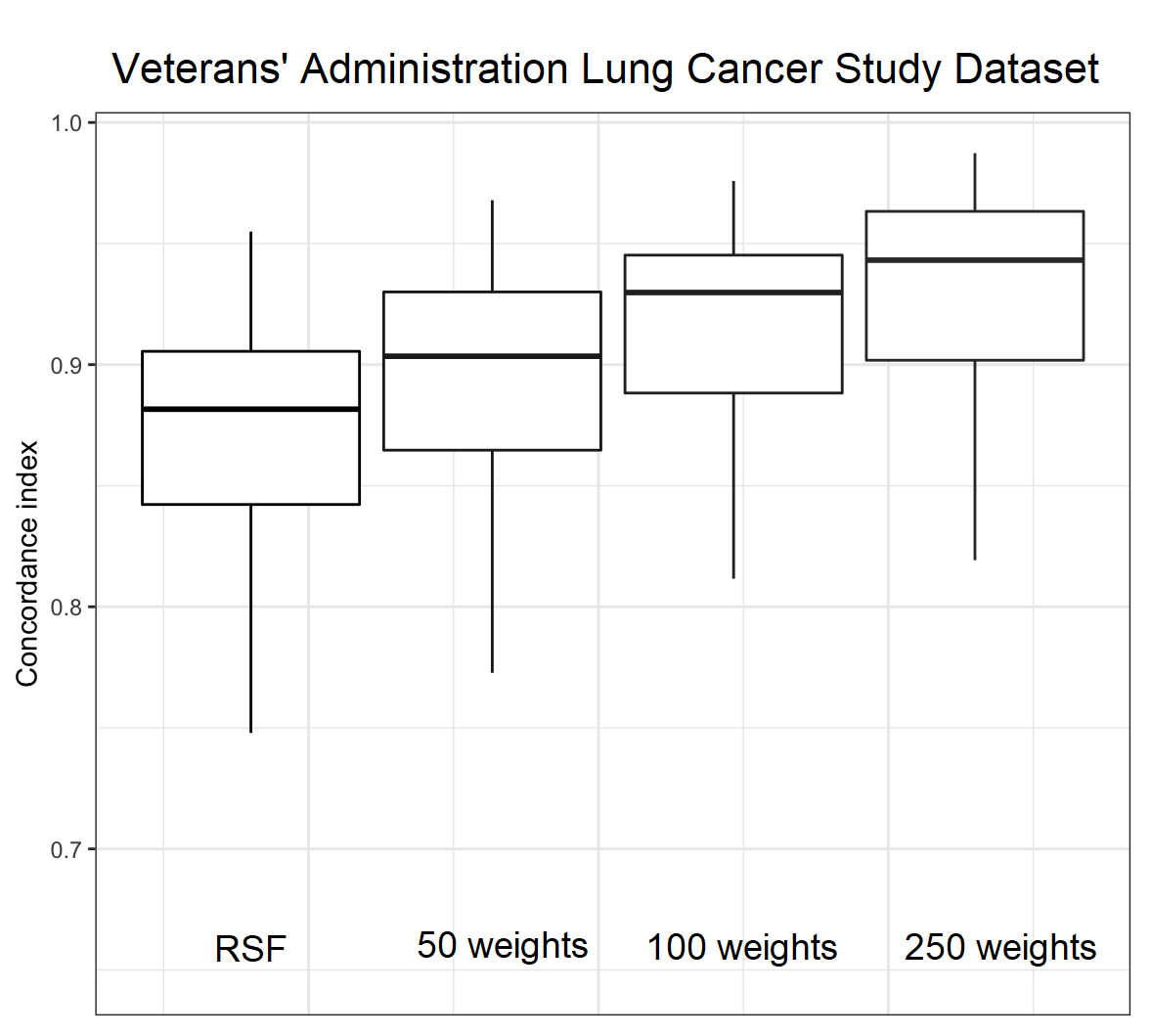}%
\caption{The boxplot for the Veteran dataset}%
\label{t:Veteran}%
\end{center}
\end{figure}

We have mentioned that in the previous section that the number of trained
weights may lead to reduction of the WRSF performance due to overfitting.
Therefore, it is interesting to study how the C-index depends on the weight
number. We take $500$ trees and divide them into $50$, $100$, $250$ subsets
such that every subset contains $10$, $5$, $2$ trees, respectively. Every
subset of trees can be viewed as a small RSF and it has its trained weight.
The corresponding boxplots of the model performances by $50$, $100$, $250$
weights for all datasets are shown in Figs. \ref{t:PBC}-\ref{t:Veteran}. It
turns out that the number of weights improves the WRSF performance for all
datasets. Moreover, it is clearly seen from the boxplots that the WRSF
outperforms the RSF especially by large number of weights.%

%

%

%

%

%

%

\begin{figure}
[ptb]
\begin{center}
\includegraphics[
height=2.1361in,
width=2.9153in
]%
{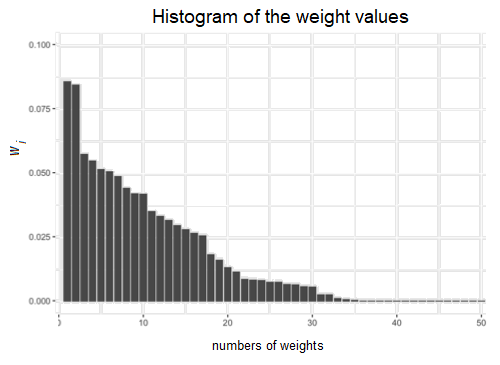}%
\caption{A histogram of the weight values}%
\label{fig:hist1}%
\end{center}
\end{figure}

An interesting question is which values of weights are assigned to trees. In
order to answer this question, we provide a typical histogram of the weight
values derived for the dataset CML (see Fig. \ref{fig:hist1}). The weights are
sorted in the descending order. The largest weight is $0.086$, the smallest
weight is $4\times10^{-4}$.%

\begin{figure}
[ptb]
\begin{center}
\includegraphics[
height=2.4984in,
width=3.147in
]%
{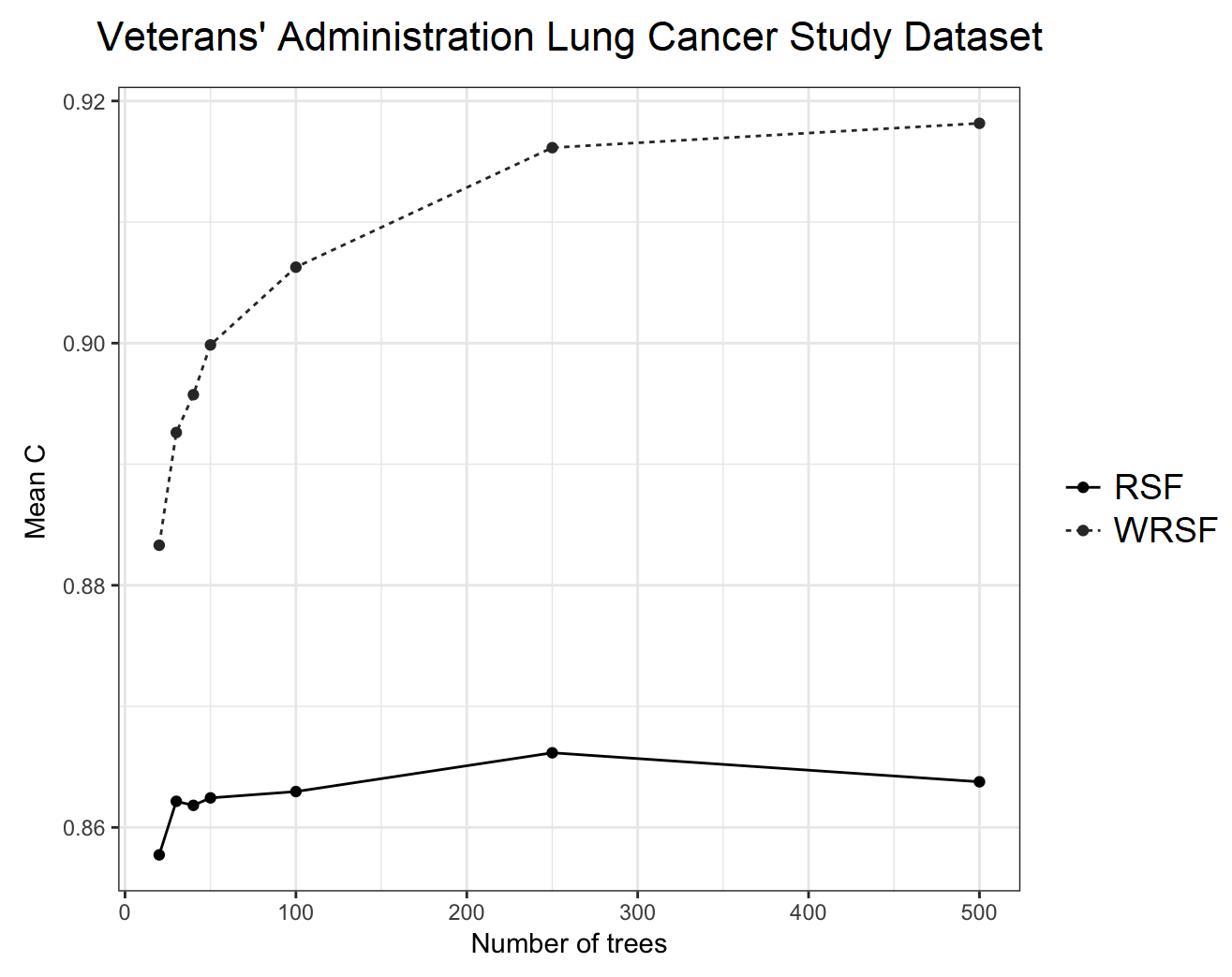}%
\caption{Dependence of the C-index on the number of trees}%
\label{fig:Veteran_tree_number}%
\end{center}
\end{figure}

Another interesting question is how the model performance depends on the
number of trees in the random forest. The dependence of the C-index on the
number of trees is illustrated in Fig. \ref{fig:Veteran_tree_number} where the
solid and dotted lines correspond to the RSF and the WRSF, respectively. It
can be seen from Fig. \ref{fig:Veteran_tree_number} that the large number of
trees may lead even to the performance deterioration when we use the RSF.
Whereas the large number of trees improves the WRSF.%

\begin{figure}
[ptb]
\begin{center}
\includegraphics[
height=2.6731in,
width=3.1306in
]%
{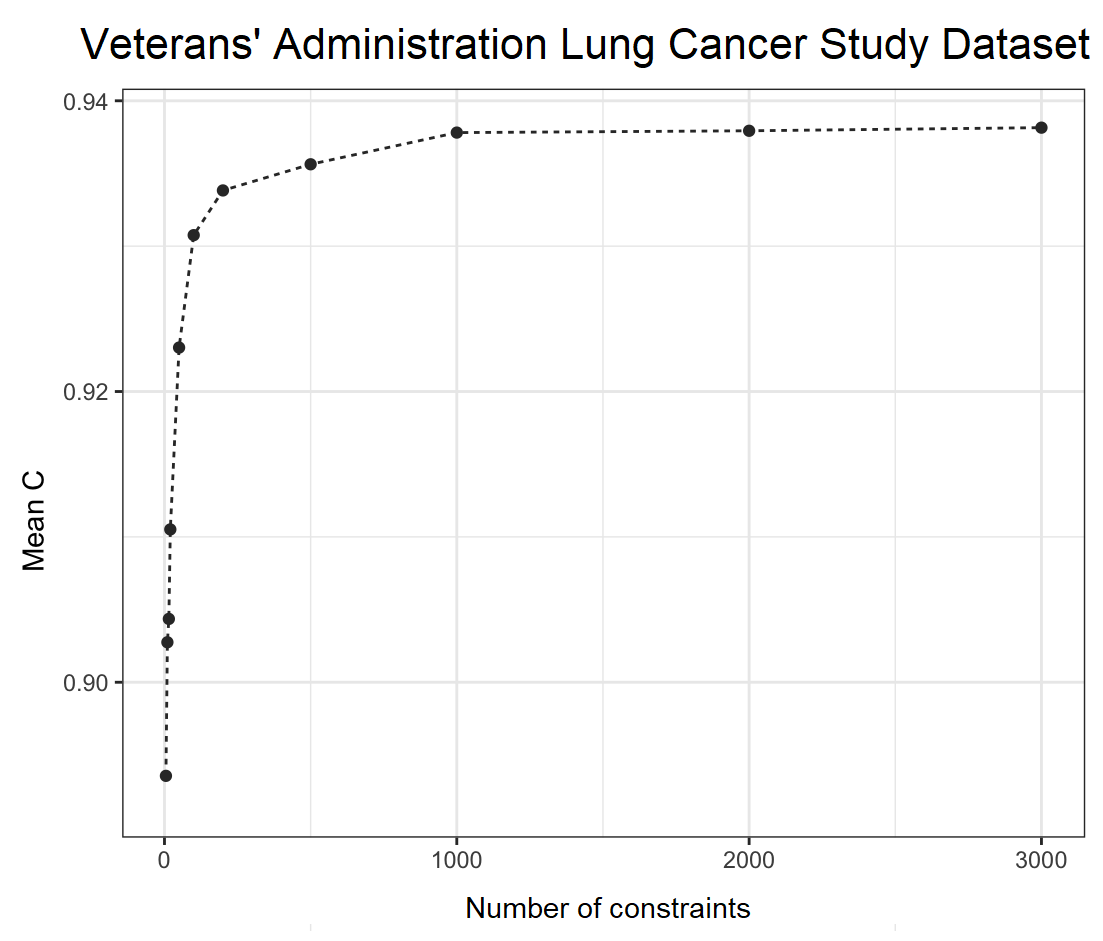}%
\caption{Dependence of the C-index on the number of selected constraints}%
\label{fig:constraints}%
\end{center}
\end{figure}

Our next experiment aims to check whether we can apply the constraint
reduction procedure in order to simplify calculations. We reduce the number of
constraints by random selection of $K$ constraints for $\xi_{ij}$ from the
whole set of constraints which is defined by all pairs of indices in the set
$J$. We use again the Veteran dataset for experiments. Fig.
\ref{fig:constraints} shows the dependence of the C-index on the number of
selected constraints for optimization. We can see from Fig.
\ref{fig:constraints} that the C-index increases with the number of
constraints. Moreover, it is important to note that the C-index for RSF is
less than the corresponding C-index for WRSF. This implies that the number of
constraints may be reduced in many cases.%

\section{Conclusion}

A new survival model based on using the weighted modification of the RSF has
been presented in the paper. The main idea underlying this model is to improve
the RSF by assigning weights to survival decision trees or to their subsets.
The weights are viewed as training parameters. It turns out that this approach
provides very improved results especially for some datasets, for example,
BLCD, LND, HTD, Veteran. Numerical experiments have illustrated that the
proposed model may provide significantly better results in comparison with the
original RSF.

The proposed model has several advantages. First, the weights are assigned in
accordance with the tree capability to correctly determine the cumulative
hazard function. Second, the weights are training parameters which are
computed by solving the standard quadratic optimization problem. As a results,
the proposed approach is very simple. But the main advantage of the model is
that it opens a door for developing a controllable RSF which can solve various
machine learning problems in the framework of survival analysis, including,
for example, transfer learning. This can be done by changing the loss function
which depends on the weights. The consideration of these problems is a
direction for further research.

We have studied only the case of linear dependence of the C-index on the
weights. However, it is interesting to consider non-linear cases. One of the
ways for implementing this case is to use a neural network which is trained to
maximize the obtained C-index. The application of the neural network as an
additional element of WRSF is also a direction for further research.

Another problem of the WRSF as well as the RSF is that the number of cases
when $\mathbf{x}_{i}$ falls into the $k$-th terminal node of a tree may be
very small. It makes confidence bounds for the Nelson--Aalen estimator, which
estimates the cumulative hazard function, to be very large. The development of
robust models taking into account this problem is another direction for
further research.

\section*{Acknowledgement}

This work is supported by the Russian Science Foundation under grant 18-11-00078.


\end{document}